\documentclass[acmsmall,screen,nonacm]{acmart}

\makeatletter
\AtBeginDocument{
	\setlength{\oddsidemargin}{\dimexpr (\paperwidth - \textwidth)/2 - 1in \relax}
	\setlength{\evensidemargin}{\oddsidemargin}
}
\makeatother



\graphicspath{{./Figures/}}
\usepackage{epstopdf}
\usepackage{amsmath,amsfonts}
\usepackage{algorithm}
\usepackage{algorithmic}
\usepackage{graphicx}
\usepackage{stfloats}

\usepackage{subfigure}
\usepackage{textcomp}
\usepackage{xcolor}
\usepackage{multicol}
\usepackage{longtable}
\usepackage{booktabs}
\usepackage{multirow}

\usepackage{longtable}
\usepackage{booktabs} 

\setcopyright{acmcopyright}
\acmDOI{XXXXXXX.XXXXXXX}

\begin{document}


\title[Graph-based Density-aware Hierarchical Clustering]{Density-aware Hierarchical Clustering Based on Element-Categorized Connection Subgraphs}


\author{Yuning Yu}
\email{yuning_yu@tongji.com}
\orcid{0000-0003-4797-8674}

\affiliation{%
	\institution{Tongji University}
	\streetaddress{Cao'an road 4800}
	\city{Shanghai}
	\country{China}
	\postcode{201804}
}

\author{Jos\'e Rodr\'iguez-Pi\~neiro}
\affiliation{%
	\institution{Tongji University}
	\streetaddress{Cao'an road 4800}
	\city{Shanghai}
	\country{China}}
\email{j.rpineiro@tongji.edu.cn}

\author{Xuefeng Yin}
\affiliation{%
	\institution{Tongji University}
	\streetaddress{Cao'an road 4800}
	\city{Shanghai}
	\country{China}}
\email{yinxuefeng@tongji.edu.cn}

\author{Bin Feng}
\affiliation{%
	\institution{RadioSky (Shanghai) Communication Technology Co., Ltd}
	\streetaddress{Cao'an road 4800}
	\city{Shanghai}
	\country{China}}
\email{fengbin@srtc.org.cn}

\renewcommand{\shortauthors}{Yu et al.}

\begin{abstract}
	Clustering is a fundamental data mining technique for pattern recognition through unsupervised learning. Among various clustering methods, hierarchical clustering,  density-based clustering, and graph clustering stand out as representative approaches. For hierarchical clustering, it can be categorized into agglomerative and divisive modes to construct clusters in a recursive manner. The key aspect of both modes is the calculation of inter-cluster similarity, which determines whether to merge the sub-clusters into one cluster or divide a current cluster into sub-clusters. Traditionally, the similarity is derived from pairwise distances, often overlooking density variations and structural connectivity in graphs. To address this, we propose a density-aware hierarchical clustering method based on element-categorized connection subgraphs (DHC-ECS)\footnote{The source code is publicly available at \url{https://github.com/yuning-yu/DHC-ECS-clustering-algorithm}.}, which effectively  integrates the hierarchical clustering, density-based clustering, and graph clustering. Particularly, a novel inter-cluster similarity metric is introduced that considers not only distances but also the element categorization in the KNN connection subgraphs, kernel density estimation, and local connectivity within sub-clusters. Extensive evaluations on heterogeneous benchmark datasets demonstrate that DHC-ECS exhibits superior overall performance in terms of clustering accuracy and parameter robustness compared with the baseline methods (including AChameleon, RNN-DBSCAN, McDPC, and G-RMS). The work indicates the great potential of the proposed clustering algorithm for low-dimensional datasets by leveraging local density and graph-structured connectivity (i.e., the duality of vertices and edges), as well as the possibility to determine an intrinsic threshold, reducing the reliance on manual parameter tuning. 
\end{abstract}





\maketitle

\section{Introduction}
A cluster is defined as a group of observations that are similar to each other, i.e. close to each other in coordinates, parameter, or attributes. The aim of clustering is to find clusters of data points with high intra-cluster similarity and low inter-cluster similarity. As a kind of unsupervised learning tool, clustering is useful for recognizing underlying patterns from large datasets, which has wide applications in Geographic Information System (GIS), image segmentation, text analysis, bioinformatics, etc \cite{8240674,6814687}.

\subsection{Background of Hierarchical Clustering}
Clustering methods can be traditionally categorized into several main types, including partitioning-based clustering, hierarchical clustering, density-based clustering, graph-based clustering, grid-based clustering, and model-based clustering. Among them, hierarchical clustering is a very common clustering method, which forms clusters recursively in an agglomerative or divisive mode. Specifically, for the agglomerative hierarchical clustering methods, initially each observation is regarded as a cluster, or sub-clusters are first generated, and then the most similar sub-clusters are merged based on the predefined and calculated similarity metric. After each agglomeration, similarity between each pair of sub-cluster is recalculated before the next agglomeration. The algorithm will finish when a certain break condition is achieved, e.g, the predefined number of clusters, or the threshold of similarity. On the contrary, the divisive hierarchical clustering method regards all observations as a cluster initially and then find the most appropriate division position according to the similarity metric. After each partition, similarity is recalculated before the next partition until the break condition is satisfied. Traditionally, the similarity can be calculated by the minimum distance, maximum distance, average distance, centroid distance \cite{1998CURE}. These simple similarity definitions have some drawbacks and cannot cope with the datasets close to each other, or without hyperspherical shapes and nonuniform sizes \cite{1998CURE}. For minimum distance and maximum distance, they are easily affected by the outlier and do not consider the internal structure of the clusters. For average distance and centroid distance, the effect of outliers also exists. To make the clustering process more robust and able to identify clusters of non-spherical shapes and wide variances in size, Cure algorithm \cite{1998CURE} represents clusters with multiple well-scattered points and shrink them toward the center of clusters, but it ignores the information of aggregate interconnectivity. Rock algorithm \cite{754967} considers the interconnectivity but it ignores the closeness of two clusters as defined by the similarity of the closest items across two clusters. Considering the unreasonable inter-cluster similarity and static model assumption, Chameleon algorithm \cite{781637} was proposed in 1999, which considers the relative interconnectivity and relative closeness. Afterwards, diverse similarity metrics are proposed including utilizing intersection points, i.e. shared nearest neighbors \cite{7439517,8596795}, similarity metric based on the modified relative interconnectivity and relative closeness considering the internal expected value for connection \cite{2019Research}, similarity metric based on adaptive neighbor graph \cite{2018An}.\par 
\subsection{Limitations of Existing Hierarchical Clustering}
\begin{figure}[!b]
	\centering
	\includegraphics[width=0.99\textwidth]{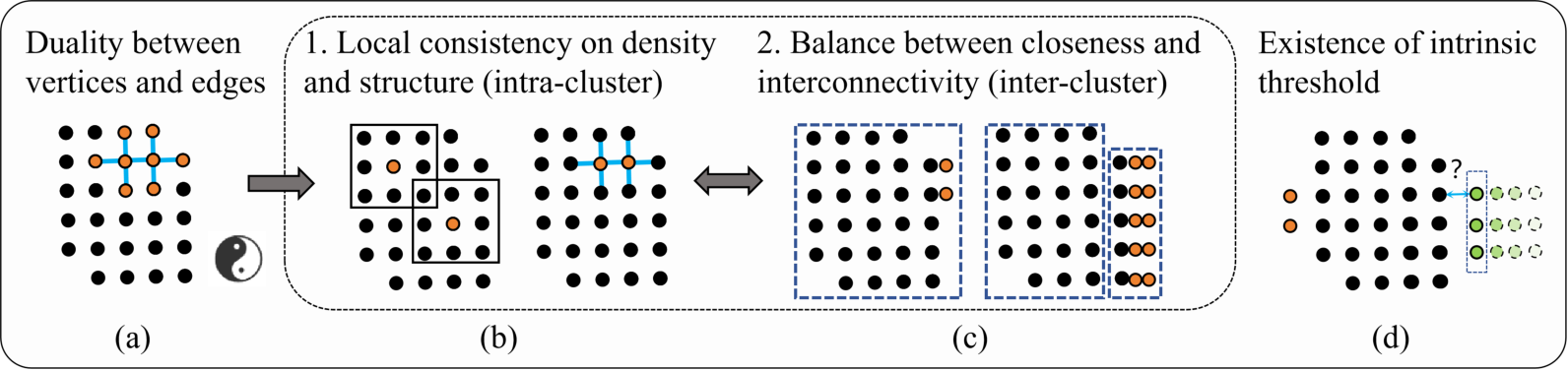} %
	\caption{Utilizable intrinsic features of clusters.}
	\label{fig:teaser}
\end{figure}

From the above investigation, the existing definitions of inter-cluster similarity for hierarchical clustering  predominantly consider the distance between sub-clusters while neglecting the density of points. For a given observation, density is defined as a measure quantifying the degree of closeness between this observation and its neighboring points. Over the years, density-based clustering algorithms have been extensively explored and have gained significant attention, leading to the development of well-known methods such as density-based spatial clustering of applications with noise (DBSCAN) \cite{ester1996density}, density peak clustering (DPC) \cite{rodriguez2014clustering,guo2024hybrid}, and mean-shift (MS) algorithm \cite{comaniciu1999mean,guo2024hybrid}. These algorithms have demonstrated their effectiveness across diverse datasets. As illustrated in Fig.~\ref{fig:teaser}(a), the duality between vertices and edges is an intrinsic property of the fully connected graph of the original data. Vertices and edges provide complementary characterizations of the underlying data structure, analogous to the concept of ``yin'' and ``yang'' in traditional Chinese philosophy. Leveraging this duality, a comprehensive clustering algorithm incorporating local structure and density is expected to yield superior clustering performance compared with independent hierarchical or density-based clustering methods, as shown in Fig.~\ref{fig:teaser}(b). In other words, the consistency in both local density and structures should be maintained within each cluster, and their variations across different clusters can be exploited for effective cluster discrimination. Meanwhile, the closeness and interconnectivity in traditional hierarchical clustering methods are also a manifestation of the duality.\par 
Existing approaches that integrate density-based and hierarchical clustering methods predominantly focus on constructing density-driven hierarchical structures \cite{campello2015hierarchical,xu2016denpehc,neto2019efficient,zhu2022hierarchical,wang2025fast}, e.g., employing density information to generate sub-clusters, and subsequently organize them into hierarchical representations, or directly exploring the evolution of clusters across different density levels to obtain multi-scale clustering hierarchies. Although these methods effectively capture clusters with arbitrary shapes and varying densities, they mainly characterize clusters from the perspective of density distribution, and graph or tree structures primarily serve as computational substrates for density-based hierarchy construction. Recent graph-based extensions \cite{shao2018graph,du2024adpscan} further combine topological properties of graphs together with hierarchical or density cluster features, providing additional information for cluster identification. Nevertheless, how to jointly exploit density consistency, structural consistency, adaptive cluster merging, and automatic hierarchy selection within a unified framework remains an open challenge.\par 

Furthermore, existing graph-based hierarchical clustering algorithms relying on graph cut criteria, such as minimum-cut and expected-cut based methods \cite{781637,shao2018graph}, primarily evaluate inter-cluster similarity according to global graph cut criteria, while the structural characteristics of the connection region between adjacent sub-clusters are only implicitly considered. Here, the connection region is defined as the set of vertices and edges linking two adjacent sub-clusters in the $k$-nearest neighbor (KNN) graph. Consequently, the estimated similarity between sub-clusters may become unreliable, particularly when they exhibit substantially different local densities. Inspired by the concepts of core, boundary, and outlier points in density-based clustering, explicitly categorizing the vertices and edges within the connection region according to their structural and density characteristics can provide a more discriminative representation of inter-cluster relationships, thereby improving clustering accuracy and robustness. This observation motivates the proposed element-categorized connection subgraph (ECS).\par

\subsection{Intrinsic Thresholds for Clustering}

Apart from the integration of hierarchical and density-based methodologies, and the refinement of vertex and edge categorization, another critical challenge in clustering lies in the selection of appropriate threshold parameters, which determines the stopping condition and the resultant clustering performance. Although some parameter-free clustering methods have been proposed \cite{hou2016dsets,li2020multiview,mahon2025k}, their effectiveness has not yet been widely validated, and their applicability to diverse clustering scenarios remains limited. Furthermore, the authors posit the existence of intrinsic and dataset-independent thresholds under a specified clustering evaluation criterion. As illustrated in Fig.~\ref{fig:teaser}(d), although the orange points can be clearly regarded as belonging to the same cluster as the black points, the cluster affiliation of the green points becomes ambiguous as they approach the black points. However, for a given clustering evaluation criterion, the continuous increase of the distance between green points and black points eventually results in a discrimination beyond a certain distance threshold. This suggests the potential existence of an inherent distance threshold determining cluster membership, which is analogous to the marginal affiliation value in fuzzy clustering theory \cite{8610270}. In practice, the relationships among sub-clusters are usually more explicit than the ambiguous case in Fig.~\ref{fig:teaser}(d), facilitating the setting of an intermediate threshold. Therefore, instead of relying on manually specified parameters or parameter-free strategies that implicitly search for optimal configurations, an alternative direction is to identify and explicitly exploit the intrinsic threshold induced by the underlying data structure, thereby enhancing generalizability of clustering methods across datasets without  explicitly defined parameters. This work provides a preliminary exploration of this idea to identify the intrinsic threshold.\par 

\subsection{Contributions of This Work}

Based on the above investigations and discussions, we propose a novel similarity metric that jointly considers both hierarchical and density characteristics of observations. The connectivity between sub-clusters is established through the ECS, where the vertices and edges within the connection regions are explicitly categorized according to their structural properties. Subsequently, the inter-cluster similarity is quantified by incorporating the structural and density information associated with these categorized connection elements. The proposed formulation comprehensively integrates link compactness (LC), link similarity (LS), density similarity (DS), and the coefficient of variation (CV) of linkage edge weights. Detailed definitions of these indices are provided in Section~\ref{sec:methodology}.\par 

The main contributions and novelties of this work include:
\begin{enumerate}
	\item \textbf{Hybrid clustering framework}: A Density-aware Hierarchical Clustering (DHC) method based on element-categorized connection subgraphs (DHC-ECS) is proposed to combine the principles of hierarchical clustering, density-based clustering, and graph-based clustering. By designing a novel inter-cluster similarity metric that explicitly considers local structure and density, DHC-ECS effectively utilizes the duality of vertices and edges in the constructed graph, enhancing clustering robustness.
	\item \textbf{Element-categorized graph representation}: To accommodate density variations and structural heterogeneity, we propose a novel KNN connection subgraph representation with explicit element categorization, namely the ECS framework, which accurately characterizes local structure within the dataset. This approach significantly improves clustering performance on datasets with inhomogeneous distributions.
	\item \textbf{Intrinsic threshold}: Our experimental results provide empirical evidence for the potential existence of the intrinsic threshold in clustering. Although the thresholds still exhibit dataset-dependent variations, the discrepancy is relatively moderate, indicating the possibility of identifying the intrinsic threshold. This work extends the potential clustering paradigm beyond conventional parameter-driven and parameter-free approaches.
\end{enumerate}

The rest of the paper is structured as follows: Section~\ref{sec:algorithm_basics} introduces the principle of hierarchical clustering and discusses the algorithm performance limitations. Section~\ref{sec:methodology} introduces the methodology of the proposed clustering algorithm, including the algorithm flow, the concept of KNN connection subgraph, the categorization of vertices and edges of the connection regions of sub-clusters, and the proposed similarity metric. In Section~\ref{sec:experiment}, we present evaluation results on ten synthetic datasets and the comparison with some baseline hierarchical and density-based clustering methods. Finally, conclusions are drawn in Section~\ref{sec:conclusion}.

\section{The Basics of Chameleon-based Hierarchical Clustering}\label{sec:algorithm_basics}
In this section, the fundamental concepts of hierarchical clustering algorithms are introduced. Particularly, a representative algorithm named Chameleon \cite{781637} is presented to facilitate a deep understanding of hierarchical clustering methods. Furthermore, the limitations of traditional Chameleon clustering methods are discussed. Several representative improvements of this algorithm are also introduced. 
\subsection{Principle of the Algorithm}
Chameleon is a well-known dynamic hierarchical clustering algorithm \cite{781637} . It models the dataset as a sparse $k$-nearest-neighbor (KNN) graph $G_{\text{knn}}(V,E)$ based on a similarity matrix computed using a predefined similarity metric between each pair of observations. Here, $V$ denotes the set of observations and $E$ denotes the set of undirected edges, satisfying that $(u,v)\in E$ if and only if $v\in N_k(u)$ or $u\in N_k(v)$, where $N_k(x)$ denotes the KNN of the observation $x$.

Based on the KNN graph, the algorithm consists of two phases. During the first phase, the KNN graph is partitioned into a set of relatively small sub-clusters. During the second phase, the sub-clusters are dynamically merged using an inter-cluster similarity considering not only relative interconnectivity but also relative closeness. The pair of sub-clusters that has the highest inter-cluster similarity is merged into one cluster and then the inter-cluster similarity between this newly generated sub-cluster and all other sub-clusters is updated before the next agglomeration. The algorithm terminates when the user-specified thresholds for relative interconnectivity and relative closeness are reached.

Particularly, the inter-cluster similarity is defined as follows. It takes relative interconnectivity $\mathrm{RI}$ and relative closeness $\mathrm{RC}$ into account comprehensively. For two sub-clusters $C_i$ and $C_j$, $\mathrm{RI}(C_i,C_j)$ is calculated as
 
\begin{equation}
	\mathrm{RI}(C_i,C_j) = \frac{{2 {\mathrm{EC}({C_i},{C_j})}}}{{ {\mathrm{EC}({C_i})} + {\mathrm{EC}({C_j})}}},
\end{equation} 
where $\mathrm{EC}(C_i,C_j)$ is the sum of the weights of the edges that straddle the two clusters, which represents the absolute interconnectivity between the two sub-clusters. $\mathrm{EC}(C_i)$ and $\mathrm{EC}(C_j)$ are the sum of the weight of the edges crossing a min-cut bisection that splits
the cluster into two roughly equal parts, which represents the internal interconnectivity of the corresponding sub-cluster. Through the normalization process, $\mathrm{RI}(C_i,C_j)$ considers the relative interconnectivity.

The relative closeness $\mathrm{RC}(C_i,C_j)$ is calculated as
\begin{equation}
	\mathrm{RC}(C_i,C_j) = \frac{{\overline{\mathrm{SEC}} ({C_i},{C_j})}}{{\frac{{\left| {{C_i}} \right|}}{{\left| {{C_i}} \right| + \left| {{C_j}} \right|}}\overline{\mathrm{SEC}}({C_i}) + \frac{{\left| {{C_j}} \right|}}{{\left| {{C_i}} \right| + \left| {{C_j}} \right|}}\overline{\mathrm{SEC}}({C_j})}},
\end{equation} 
where $\overline{\mathrm{SEC}}(C_i, C_j)$ is the average weight of the edges that connect vertices in $C_i$ and $C_j$, $\overline{\mathrm{SEC}}(C_i)$ and $\overline{\mathrm{SEC}}(C_j)$ are respectively the average weights of the edges that belong to the min-cut bisector of $C_i$ and $C_j$. $|C_i|$ and $|C_j|$ denote the number of points in each cluster. Inter-cluster similarity can be defined as 
\begin{equation}
	\mathrm{RI}(C_i,C_j) \times \mathrm{RC}(C_i,C_j)^\alpha
\end{equation}
where $\alpha$ is a parameter giving different importance to $\mathrm{RI}(C_i,C_j)$ and $\mathrm{RC}(C_i,C_j)$.\par

\subsection{Discussion on Algorithm Performance}

By dynamically updating the inter-cluster similarity considering both relative interconnectivity and relative closeness, Chameleon algorithm balances inter-cluster and intra-cluster characteristics without relying on static model assumptions. Therefore, it can obtain better clustering results on datasets of different shapes, sizes, and densities compared with traditional hierarchical methods \cite{781637}. However, some drawbacks of the Chameleon algorithm and other hierarchical clustering methods can be identified, as demonstrated in the introduction part, including: (1) insufficient consideration of local characteristics within and near the connection regions between sub-clusters; (2) lack of density and graph structural information for the similarity calculation; (3) threshold dependency.\par 

For the past few decades, several improvements to the Chameleon algorithm have been proposed. Existing studies mainly focus on refining the similarity measure \cite{guo2019research,barton2019chameleon}, incorporating local density and adaptive neighborhood graphs \cite{2018An}, improving the robustness of hierarchical merging and automatic cutoff selection \cite{jeong2013data,barton2019chameleon}, and enhancing computational efficiency for large-scale datasets \cite{singh2025chameleon2++}. Despite these improvements, the underlying graph representation remains largely unchanged. Consequently, the structural characteristics of the connection region between neighboring sub-clusters are still inadequately characterized, while both graph construction and hierarchical merging continue to rely heavily on manually specified thresholds. Furthermore, existing similarity measures mainly evaluate the overall interconnectivity or closeness between sub-clusters, without explicitly exploiting the structural properties of individual vertices and edges within the connection region. These limitations motivate the proposed DHC-ECS algorithm, which explicitly models both structural and density information for inter-cluster similarity estimation, based on the proposed element-categorized KNN connection subgraphs.

\section{The Proposed Clustering Method}\label{sec:methodology}

The proposed DHC-ECS algorithm can be divided into two stages. In the first stage, a set of relatively small sub-clusters is generated, and preliminarily merged in a greedy way by iteratively agglomerating the pair of sub-clusters with the highest similarity. Since the sub-clusters in this stage are relatively small, their similarity is measured using the average distance between points. In the second stage, since the sub-clusters become larger and contain more points, the average similarity is not sufficient for characterizing inter-cluster similarities. Therefore, the proposed inter-cluster similarity metric is applied. The detailed algorithm flow is described below. 

\begin{enumerate}
	\item {\textbf{Greedy merging}}: In the first stage, each observation is initially regarded as a separate cluster, and the sub-clusters are then merged iteratively through a greedy method \cite{2019Research}, i.e., we find the pair of sub-clusters that has largest similarity and merge them into one cluster. Then the similarity between this new cluster and the other clusters is updated. In this stage, the similarity is calculated based on the average distance between points in different clusters as
	\begin{align}
		\label{eq:weight_similarity}
		{d_{i,j}} &= \frac{1}{{\left| {C_i}\right| \left| {C_j}\right|}}\sum\limits_{q \in {C_j}} {\sum\limits_{p \in {C_i}} {{\rm dist}(p,q)}},\\
		{w_{i,j}} &= \frac{1}{{{d_{i,j}} + 1}}, \quad S_{i,j}=w_{i,j}.
	\end{align}
	Here, $d_{i,j}$ denotes the distance between $C_i$ and $C_j$, where $C_i$ and $C_j$ are the $i$-th and $j$-th cluster, respectively. ${\rm dist}(p,q)$ represents the distance between point pair $p$ and $q$, where $p$ and $q$ belong to $C_i$ and $C_j$, respectively. This distance can be defined as needed, e.g., geodesic distance for non-convex clusters, standard Euclidean distance for data with inconsistent scales across dimensions, or Mahalanobis distance when correlation exists between dimensions. $\left|.\right|$ denotes the number of observations in a cluster. $w_{i,j}$ denotes the edge weight, which also serves as the similarity $s_{i,j}$ between $C_i$ and $C_j$. For the synthetic datasets in Section~\ref{sec:experiment}, 
	we adopt geodesic distance rather than Euclidean distance, because the dimensions are balanced and independent while manifold structures are present. This allows the distance measure to better reflect the actual distance on the manifold. The calculation process of geodesic distance is introduced below, where KNN is adopted for the calculation.
	
	\begin{enumerate}
		\item Calculate the Euclidean distance between all pairs of points and find the KNN set $N_{k}(x)$ of each point $x$ in the whole dataset $\mathcal X$.
		\item Initialize the geodesic distance. 
		\begin{equation}
			{d_{\rm{G}}}(p,q) = \left\{ \begin{array}{l}
				d(p,q),{\rm{if\:}}q \in {N_{k}}(p)\\
				\infty ,{\rm{otherwise}}	
			\end{array} \right.
		\end{equation}
		\begin{equation}
			{d_{\rm{G}}}(p,q) = {\rm min}\{{d_{\rm{G}}}(p,q),{d_{\rm{G}}}(q,p)\},
		\end{equation}
		Here, $d_G(p,q)$ represents the geodesic distance between $p$ and $q$, and $N_{k}(p)$ represents the KNN set of $p$.
		\item Calculate the shortest distance between any pair of points $p$ and $q$.
		\begin{equation}
			{d_{\rm{G}}}(p,q) = \mathop {\rm min}\limits_{h \in {\mathcal X} \setminus \{p,q\}}\{{d_{\rm{G}}}(p,q),{d_{\rm{G}}}(p,h)+{d_{\rm{G}}}(h,q)\},
		\end{equation}
		where $h$ is an intermediate point besides $p$ and $q$. The shortest distance between any pair of point is obtained as the final geodesic distance.
		
	\end{enumerate}
	
	The stopping condition for the first stage is that the number of sub-clusters equals the threshold $T_\text{n}$, which is larger than the actual number of clusters. It is commonly set to $\sqrt{N}$ in \cite{2019Research}, where $N$ denotes the number of points in the entire dataset.	
	
	\item {\textbf{Refined merging}}: In the second stage, an agglomerative procedure is conducted to obtain the final clustering results, while the similarity calculation changes since the sub-clusters are larger in size than that in the first stage and the local characteristic becomes more important to determine whether to merge two clusters. For this reason, we construct a KNN graph for the entire dataset and determine the linkage between clusters. We use $N_{k(x)}(x)$ to denote the KNN set\footnote{It should be noted that if the total number of points in a sub-cluster is less than the preset $K$, the $K$ value for this sub-cluster can be set to be the number of its points, and the KNN connection subgraph may become asymmetric for this sub-cluster and its connected sub-clusters. This processing ensures the balance of internal elements and linkage elements for small sub-clusters.} of point $x$ , and use $E_{k(x)}(x)$ to denote the set of edges between point $x$ and its KNN. More generally, for any point set  $\mathcal{X}$, we define ${\mathcal{N}_{k(\cdot)}}(\mathcal{X})= \bigcup_{x \in \mathcal{X}} N_{k(x)}(x)$, $\mathcal{E}_{k(\cdot)}(\mathcal{X})= \bigcup_{x \in \mathcal{X}} E_{k(x)}(x)$. Assuming that the index set of cluster pairs with connections is $\{\{i_1,j_1\},\{i_2,j_2\},...,\{i_{N^t},j_{N^t}\}\} $, where $N^t$ denotes the number of cluster pairs in the $t$-th iteration, several key definitions for cluster pairs $C_i$ and $C_j$ are introduced as follows:
	\begin{itemize}
		\item \textit{\textbf{KNN graph}}\par		
		First, based on the concept of KNN introduced above, we can construct a KNN graph for the original dataset. A clear definition is written as
		\begin{equation}
			G^{\text{KNN}}=(V,E^{\text{KNN}}), V={\mathcal X},
		\end{equation}
		where 
		\begin{equation}
			 E^{\text{KNN}}=\big\{\{u,v\} \big| u\in N_{k(v)}(v) \;\ \mathrm{or} \;\  v \in N_{k(u)}(u)\big\}.
		\end{equation}
		\item \textit{\textbf{KNN connection subgraph}}\par	
	    Based on the definition of KNN graph, the KNN connection subgraph can be defined to describe the connection between a pair of clusters $\{C_i, C_j\}$. Here, $\{.,.\}$ denotes an unordered pair.  $C_i$ and $C_j$ are both point sets, and $C_i,C_j\subseteq V$. The corresponding KNN connection subgraph is defined as a subgraph constructed from the vertices in $\{C_i, C_j\}$ involved in inter-cluster KNN linkage and their KNN neighbors, denoted as
		\begin{equation}
			{G}_{i,j}^{\text{KNN}}=(V_{i,j}^{\text{KNN}},E_{i,j}^{\text{KNN}}),
		\end{equation}
		To obtain ${G}_{i,j}^{\text{KNN}}$, the vertices that generate the linkage relationship are first picked out:
		\begin{equation}
			\qquad	V_{i,j}^{*}
			=\Big\{u \,\Big|\,
			\big(u\in C_i \land \exists v\in N_{k(u)}(u)\cap C_j\big)
			\, \lor \,
			\big(u\in C_j \land \exists v\in N_{k(u)}(u)\cap C_i\big)
			\Big\},
		\end{equation}
		Then, the edges connected to these seed vertices form the edge set of ${G}_{i,j}^{\text{KNN}}$
		\begin{equation}
			E_{i,j}^{\text{KNN}}
			= \mathcal{E}_{k(\cdot)}(V_{i,j}^{*}).
		\end{equation}	
		
		Therefore, the vertex set of ${G}_{i,j}^{\text{KNN}}$ can be determined including both seed vertices $V_{i,j}^{*}$ and their KNN, formulated as 

		\begin{equation}
			V_{i,j}^{\text{KNN}} = V_{i,j}^{*} \cup \mathcal{N}_{k(\cdot)}(V_{i,j}^{*}).
		\end{equation}
		
		For an intuitive presentation, an example of a KNN connection subgraph is shown in Fig.~\ref{fig:AKNN_connection_graph}. Two neighboring sub-clusters are marked using green points as Sub-cluster~1 and blue points as Sub-cluster~2. For each point, the KNN set is determined according to the above definition, and the KNN connection subgraph is plotted as shown in Fig.~\ref{fig:AKNN_connection_graph}. According to the linkage relationship, the KNN connection subgraph can be divided into different parts. For the vertex set, it can be divided into linkage points and internal points, which are marked respectively with circles and squares. For the edge set, it can be divided into linkage edges and internal edges, which are plotted using red and black lines respectively. The detailed definitions are introduced below.
		
		\begin{figure}
			\centering
			{\includegraphics[width=0.5\columnwidth]{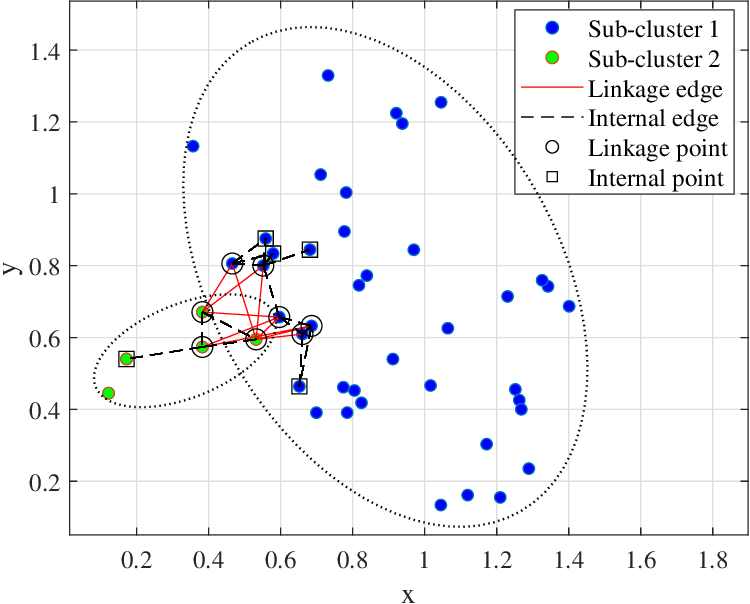}}
			\caption{An example of KNN connection subgraph and categorization of vertices and edges (k=4).}
			\label{fig:AKNN_connection_graph}
		\end{figure}
		
		\item \textit{\textbf{Linkage points}}\par
		Linkage points denote the set of vertices connecting the two neighboring sub-clusters, defined as $V_{i,j}^{\text{linkage}} = V_i\cup V_j$, satisfying 1) $V_i \subseteq C_i, V_j \subseteq C_j$, 2) $\forall v_i \in V_i$, $\exists\,v\in C_j$, s.t. $v\in N_{k(v_i)}(v_i)$ or $v_i \in N_{k(v)}(v)$. This constraint also applies to $V_j$. In Fig.~\ref{fig:AKNN_connection_graph}, linkage points refer to those points linked by red edges.
		\item \textit{\textbf{Linkage edges}}\par
		Linkage edges denote the set of edges $E_{i,j}^\text{linkage}$ connecting $V_i$ and $V_j$ in $V_{i,j}^{\text{linkage}}$, satisfying 1) $E_{i,j}^\text{linkage} \subseteq \mathcal{E}_{k(\cdot)}(V_i) \cup \mathcal{E}_{k(\cdot)}(V_j)$, 2) 
		$\forall e=\{v_1,v_2\}$, s.t. ${(v_1 \in V_i \land v_2 \in V_j)} \vee {(v_1 \in V_j \land v_2 \in V_i)}$.	The weight of each linkage edge is denoted as $W(e_{i,j}^\text{linkage})$, which can be calculated as \eqref{eq:weight_similarity}. In Fig.~\ref{fig:AKNN_connection_graph}, linkage edges refer to the red edges. 
		\item \textit{\textbf{Internal points connecting to linkage points}}\par
		Internal points denote the set of points connected to linkage points but not contributing to the KNN connection between the two sub-clusters, defined as $V_{i,j}^\text{internal}=(\mathcal{N}_{k(\cdot)}(V_i) \cup \mathcal{N}_{k(\cdot)}(V_j)) \setminus V_{i,j}^\text{linkage}$, where `$\backslash$' denotes the set-difference operator.
		\item \textit{\textbf{Internal edges connecting to linkage points}}\par
		Internal edges denote the set of edges that connect the linkage points and internal points, defined as $E_{i,j}^\text{internal}=(\mathcal{E}_k(V_i) \cup \mathcal{E}_k(V_j))\setminus E_{i,j}^\text{linkage}$, i.e, the edges that belong to the linkage points and connect to the KNN set, but are not part of the linkage edge set.
		The weight of each internal edge is denoted as $W(e_{i,j}^\text{internal})$. In Fig.~\ref{fig:AKNN_connection_graph}, internal edges connecting to linkage points refer to the black edges. 
		\item \textit{\textbf{Point Density}}\par
		Point density is defined as the kernel density used to represent the concentration degree of neighboring points. For a specific observation $x$, its density is calculated as $D(x)=\sum\limits_{y \in N_{k(x)}(x)} {{\exp{(-\left| {\text{dist}(x,y)} \right|)}}}$. 
	\end{itemize}
Based on the element categorization of the KNN connection subgraphs, the similarity between two sub-clusters is defined as 
\begin{equation}
	\label{eq:similarity}
	S_{i,j}=2\text{LC}\frac{1}{{{{\left( {\left| {\text{LS} - 1} \right| + 1} \right)}^\alpha }}}\frac{1}{{{{\left( {\left| {\text{DS} - 1} \right| + 1} \right)}^\alpha }}},
\end{equation}
where $S_{i,j}$ is the similarity metric between $C_i$ and $C_j$. It considers the link compactness (LC), link similarity (LS), density similarity (DS), and the coefficient of variation (CV) of the linkage-edge weights for automatically adjusting the similarity according to the randomness in the distribution of linkage points. The definitions of LC, LS, DS, CV, and $\alpha$ are given as follows: 
\begin{equation}
	\label{eq:LC}
	\text{LC}=\frac{{\sum\limits_{e_{i,j}^{{\rm{linkage}}} \in E_{i,j}^{{\rm{linkage}}}} {W(e_{i,j}^{{\rm{linkage}}})} }}{{\sum\limits_{e_{i,j}^{{\rm{internal}}} \in E_{i,j}^{{\rm{internal}}}} {W(e_{i,j}^{{\rm{internal}}})} }},
\end{equation}
\begin{equation}
	\label{eq:LS}
	\text{LS}=\frac{{\left| {E_{i,j}^{{\rm{internal}}}} \right|\sum\limits_{e_{i,j}^{{\rm{linkage}}} \in E_{i,j}^{{\rm{linkage}}}} {W(e_{i,j}^{{\rm{linkage}}})} }}{{\left| {E_{i,j}^{{\rm{linkage}}}} \right|\sum\limits_{e_{i,j}^{{\rm{internal}}} \in E_{i,j}^{{\rm{internal}}}} {W(e_{i,j}^{{\rm{internal}}})} }},
\end{equation}
\begin{equation}
	\label{eq:DS}
	\text{DS}=\frac{{\left| {{{V_{i,j}^\text{internal}}}} \right|\sum\limits_{v_{i,j}^\text{linkage} \in {V_{i,j}^{{\rm{linkage}}}}} {D({v_{i,j}^\text{linkage}})} }}{{\left| {{{V_{i,j}^\text{linkage}}}} \right|\sum\limits_{v_{i,j}^\text{internal} \in {V_{i,j}^{{\rm{internal}}}}} {D({v_{i,j}^\text{internal}})} }},
\end{equation}
\begin{equation}
	\label{eq:CV}
	\text{CV}=\frac{{\text{std}\left( {W\left( {e_{i,j}^{{\rm{linkage}}} \cup e_{i,j}^{{\rm{internal}}}} \right)} \right)}}{{\text{mean}\left( {W\left( {e_{i,j}^{{\rm{linkage}}} \cup e_{i,j}^{{\rm{internal}}}} \right)} \right)}},\,\alpha=e^{-r\text{CV}}.
\end{equation}
Here, $\text{std}(.)$ and $\text{mean}(.)$ denote the standard deviation and mean value of the elements in the set, respectively. $|\cdot|$ denotes the cardinality of a set. $r$ is an adjustable parameter to control the influence of the randomness of point distribution on the  similarity calculation, which is fixed to $3$ in this work. As shown in \eqref{eq:LC}--\eqref{eq:CV}, LC represents the linkage compactness between two sub-clusters as the ratio between the sum of weights of linkage edges and internal edges, LS represents the linkage similarity between two sub-clusters as the ratio between the average weights of linkage edges and internal edges. DS represents the density similarity as the ratio between average density of linkage points and average density of internal points connecting to linkage points. $\alpha$ is a coefficient used to adapt to the randomness of point distribution. When the variation of edge weights and point density is large, corresponding to a large CV, the similarity decreases automatically as $\alpha$ approaches 0, and the terms ${\left( {\left| {\text{LS}-1} \right| + 1} \right)}^\alpha$ and ${\left( {\left| {\text{DS}-1} \right| + 1} \right)}^\alpha$ are driven to approach 1 because of the exponent $\alpha$. In this case, the link compactness $\text{LC}$ becomes more significant. Additionally, according to \eqref{eq:similarity}, a larger value of $S_{i,j}$ indicates higher similarity between $C_i$ and $C_j$. The break condition for this algorithm is that the similarity for any pair of clusters is lower than the predefined threshold $T_\text{s}$, or the number of clusters equals the predefined value. 

\end{enumerate}

\begin{algorithm}[!t] 
	\small
	\setlength{\baselineskip}{\baselineskip}  
	\caption{Density-aware hierarchical clustering based on element-categorized connection subgraphs}  
	\label{alg1} 
	\begin{algorithmic}[1] 
		\REQUIRE 
		Data set: $\mathcal{X}$, Predefined parameters: $T_\text{n}$, $T_\text{s}$ (optional: $T_\text{n,final}$), $K,r$. 
		\ENSURE Clustering results: $L(x)$, Cluster number: $N_\mathrm{c}$.
		\STATE $\text{label}=1,N_\mathrm{c}=0$
		\FOR{all $x \in \mathcal{X}$}
		\STATE $L(x)\leftarrow \text{label} $
		\STATE $\text{label}\leftarrow \text{label}+1$
		\STATE $N_\mathrm{c}\leftarrow N_\mathrm{c}+1$
		\ENDFOR
		\FOR{$i=0$ to $N_\mathrm{c}-1$}
		\FOR{$j=i$ to $N_\mathrm{c}$}
		\STATE calculate $S_{i,j}$
		\STATE $S_{j,i}\leftarrow S_{i,j}$
		\ENDFOR
		\ENDFOR		
		\WHILE{$N_\mathrm{c}>T_\text{n}$}
		\STATE $\{i,j\}\leftarrow \arg \mathop {\max }\limits_{i,j} {S_{i,j}},i<j$
		\STATE $L({x_j}){\big|_{L({x_j}) = j}} \leftarrow i$
		\STATE $L({x_j}){\big|_{L({x_j}) >j}} \leftarrow L(x_j)-1$
		\STATE Update similarity matrix $\boldsymbol{S}$
		\ENDWHILE
		\STATE Construct KNN connection subgraphs
		\STATE Obtain $D(x)$, $V_{i,j}^{\text{linkage}}$, $E_{i,j}^{\text{linkage}}$, $V_{i,j}^{\text{internal}}$, $E_{i,j}^{\text{internal}}$		
		\STATE Calculate inter-cluster similarity $S_{i,j}$ for each pair of clusters 
		\WHILE{$\exists S_{i,j}>T_\text{s}$ (optional: $N_\mathrm{c} > T_\text{\textit{n},final}$) }
		\STATE $\{i,j\}\leftarrow \arg \mathop {\max }\limits_{i,j,i \ne j} {S_{i,j}},i<j$
		\STATE $L({x_j}){\big|_{L({x_j}) = j}} \leftarrow i$
		\STATE $L({x_j}){\big|_{L({x_j}) >j}} \leftarrow L(x_j)-1$
		\STATE Update similarity matrix $V_{i,j}^{\text{linkage}},E_{i,j}^{\text{linkage}}, \boldsymbol{S}$
		\ENDWHILE	
	\end{algorithmic}
\end{algorithm}

Based on the interpretation above, the proposed algorithm differs from traditional methods in three key aspects: (1) providing a more detailed description of the KNN connection subgraph structure through element categorization to separate vertices and edges into internal and interconnected components, representing intra-cluster and inter-cluster similarity respectively; (2) incorporating density and randomness information into similarity estimation; and (3) focusing on boundary connection regions of sub-clusters to achieve a more refined characterization. Consequently, the proposed algorithm is expected to achieve enhanced robustness and adaptivity.

\section{Experimental Results and Discussion}\label{sec:experiment}
\subsection{Clustering Performance on Synthetic Datasets}
\begin{figure}[!h]
	\vspace{-0mm}
	\setlength{\subfigcapskip}{-0pt}
	\centering
	\subfigure[]{\includegraphics[width=0.18\columnwidth]{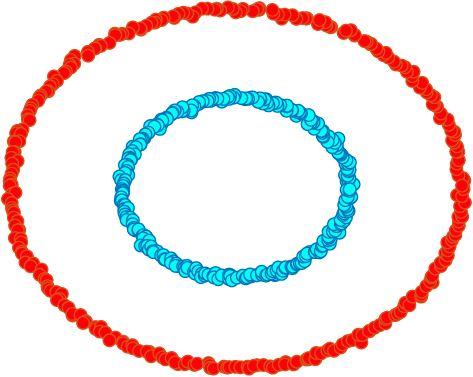}%
		\label{fig:KNN}}
	\
	\subfigure[]{\includegraphics[width=0.18\columnwidth]{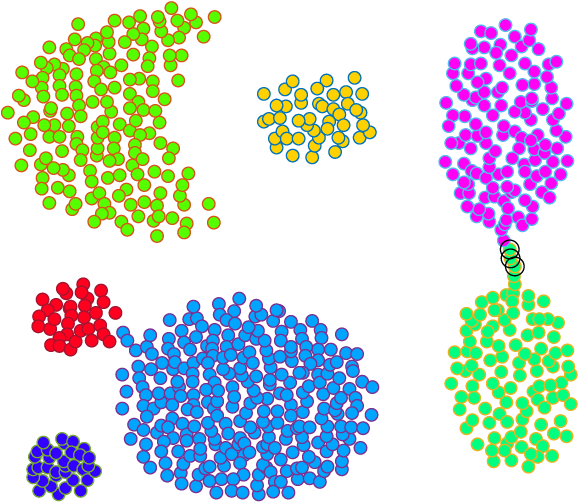}%
		\label{fig:AKNN}}
	\
	\subfigure[]{\includegraphics[width=0.18\columnwidth]{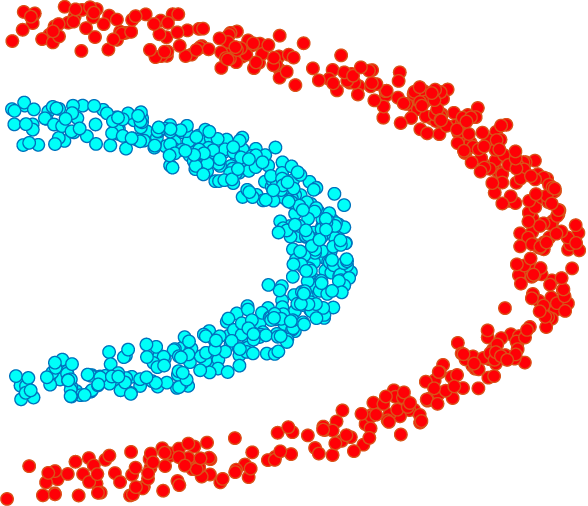}%
		\label{fig:AKNN}}
	\
	\subfigure[]{\includegraphics[width=0.18\columnwidth]{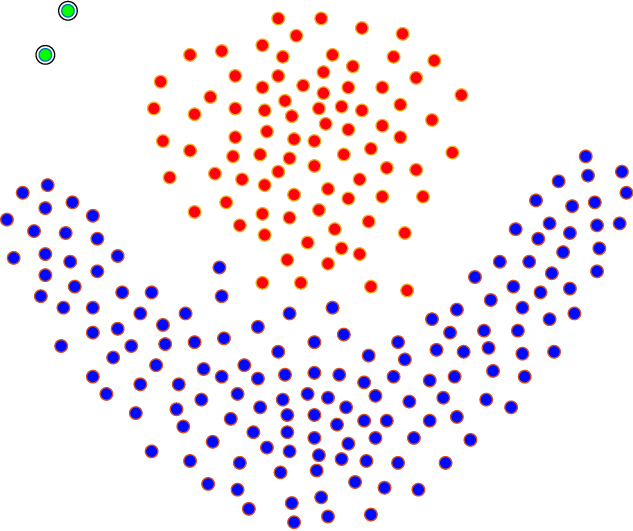}%
		\label{fig:AKNN}}
	\
	\subfigure[]{\includegraphics[width=0.18\columnwidth]{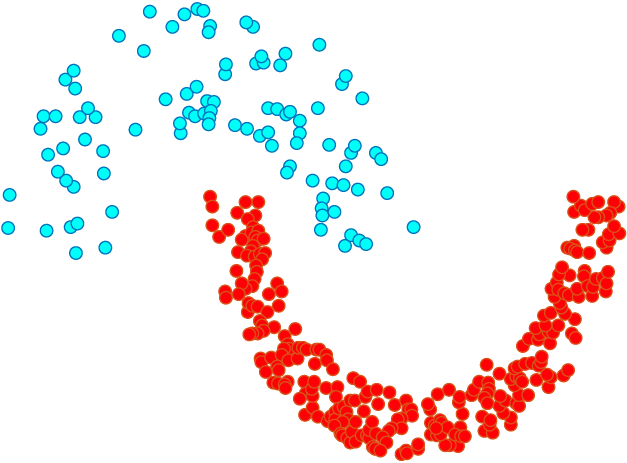}%
		\label{fig:AKNN}}
	\
	\subfigure[]{\includegraphics[width=0.18\columnwidth]{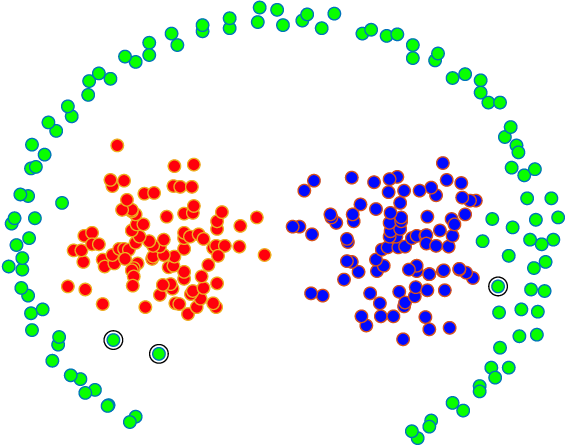}%
		\label{fig:AKNN}}
	\
	\subfigure[]{\includegraphics[width=0.18\columnwidth]{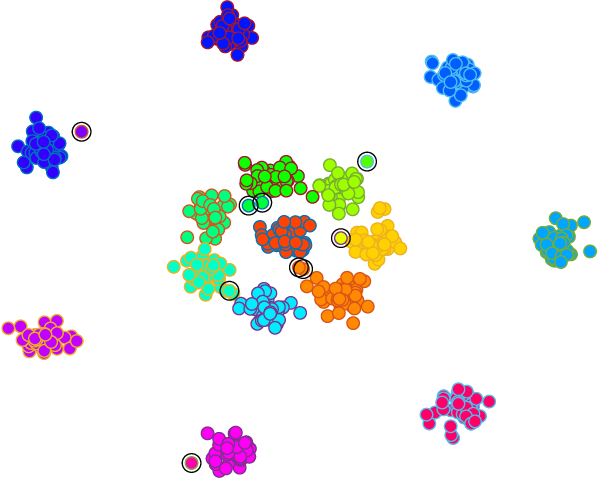}%
		\label{fig:AKNN}}
	\
	\subfigure[]{\includegraphics[width=0.18\columnwidth]{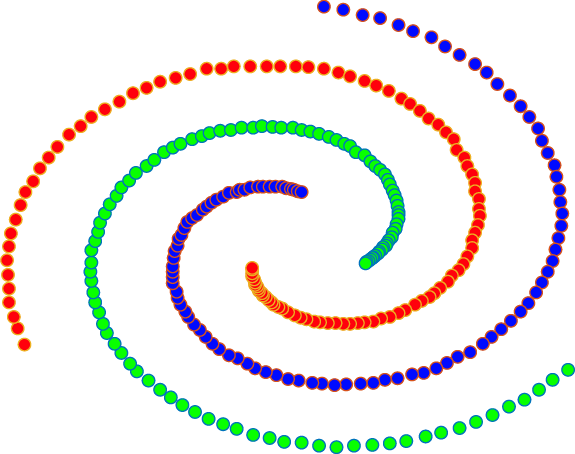}%
		\label{fig:AKNN}}
	\
	\subfigure[]{\includegraphics[width=0.18\columnwidth]{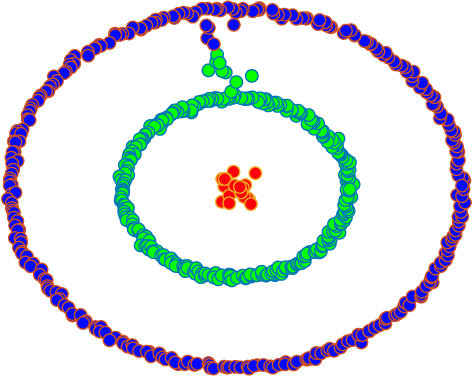}%
		\label{fig:AKNN}}
	\
	\subfigure[]{\includegraphics[width=0.18\columnwidth]{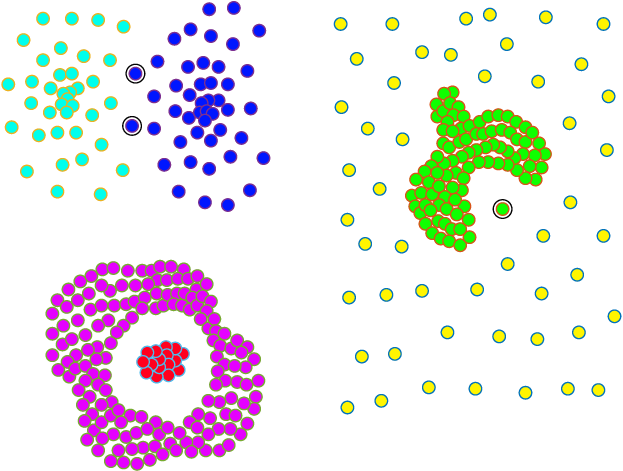}%
		\label{fig:AKNN}}
	\caption{Clustering results on synthetic datasets (The points that are wrongly categorized are circled in black). (a) 2 circles. (b) Aggregation. (c) Halfkernel. (d) Flame. (e) Jain. (f) Pathbased. (g) R15. (h) Spiral. (i) Two circles noise. (j) Compound.}
	\label{fig:clustering_results}
	\vspace{-0mm}
\end{figure}

For evaluation of the proposed clustering algorithm, ten synthetic datasets are used including 2circles, Halfkernel, Aggregation, Flame, Jain, Pathbased, R15, Spiral, Two circle noise, and Compound \cite{ClusteringDatasets,2020MATLAB}. The proposed DHC-ECS algorithm and the hierarchical algorithm with traditional similarity metrics (minimum distance, maximum distance, centroid distance, mean distance \cite{1998CURE}), as well as several recently proposed and representative algorithms, including an improved Chameleon algorithm (ACHAMELEON) \cite{2019Research}, reverse-nearest-neighbor-based DBSCAN (RNN-DBSCAN)\cite{8240674}, multi-center clustering by fast search and find of density peak (McDPC) \cite{2020MATLAB}, Gaussian kernel based robust mean-shift clustering (G-RMS)\cite{9698033}, are tested on these datasets. The clustering results of our proposed method are illustrated in Fig.~\ref{fig:clustering_results}, where different colors represent distinct detected clusters, and misclassified points are marked by black circles. We can see that across all datasets, the majority of points are correctly clustered, except for a few points with ambiguous affiliation in the KNN connection regions of sub-clusters. Some external evaluation indices including Normalized Mutual Information (NMI) \cite{2008Introduction}, Adjusted Rand Index (ARI) \cite{0Information}, Fowlkes and Mallows index (FMI) \cite{2011Powers} and Purity \cite{2008Introduction} are used to quantify the performance. To ensure a fair comparison, the Euclidean distance or the geodesic distance is adopted for all baseline algorithms other than G-RMS, depending on the dataset features and clustering performance. The evaluation results are listed in Table~\ref{table:evaluation_index}, with corresponding parameters that are applied for each dataset being listed in Table~\ref{table:parameters_applied}.\par 

From these evaluation indices and the applied parameters, we can conclude that 
\begin{enumerate}
	\item Hierarchical clustering method based on traditional similarity metrics (minimum distance, maximum distance, centroid distance, mean distance) performs quite unstably on these datasets, due to the different features of data distributions.  
	\item Apart from Spiral, Pathbased, and Compound, the ACHAMELEON algorithm performs well on most datasets with NMI larger than 0.97.
	\item RNN-DBSCAN and McDPC perform well on most datasets except Halfkernel for RNN-DBSCAN, and Compound for McDPC. In addition, the parameter tuning for these two algorithms is heuristic \cite{8240674,2020MATLAB} and time-consuming to obtain optimal results. Moreover, since the McDPC algorithm relies on the decision graph defined by the local density and the $\delta$ distance \cite{2020MATLAB}, the experiments indicate that representative points belonging to different clusters but exhibiting similar decision features may be wrongly classified to the same clusters. Besides, the G-RMS algorithm is applicable to all datasets except single-center datasets (e.g., 2circles) or nested datasets (e.g., Compound).
	\item Although the clustering results of the proposed DHC-ECS algorithm are not the best for all datasets, it exhibits better robustness and obtains good results for all these datasets (e.g, the NMI values for all the 10 datasets are larger than 0.95). Furthermore, although these datasets exhibit distinct features, the parameter applied to all the datasets are quite similar as shown in Table~\ref{table:parameters_applied}, except the threshold of similarity for stop condition (the mean value of $T_\text{s}$ is around 0.38 and the variation is moderate). Meanwhile, $T_\text{s}$ has an intuitive interpretation corresponding to the clustering granularity---a smaller threshold allows more sub-clusters to be merged, resulting in coarser clustering results, whereas a larger threshold produces finer partitions. This facilitates the parameter settings. To summarize, these results provide empirical evidence for the existence of intrinsic thresholds.
\end{enumerate}

In conclusion, the proposed DHC-ECS algorithm achieves more robust performance across these diverse synthetic datasets, compared with traditional and improved hierarchical clustering methods, as well as several recently proposed density-based clustering approaches. However, it is worth noting that due to the influence of the ``curse of dimensionality''\cite{beyer1999nearest}, both distance and density estimations, as well as structural information should be re-evaluated in the high-dimensional spaces. Therefore, further investigation is required to evaluate the effectiveness of the DHC-ECS algorithm when applied to high-dimensional datasets in the future.

{
\footnotesize
\begin{longtable}{llllll}
	\caption{Performance Evaluation on Synthetic Datasets.}
	\label{table:evaluation_index} \\
	\toprule
	\textbf{Dataset} & \textbf{Method} & \textbf{NMI} & \textbf{ARI} & \textbf{FMI} & \textbf{Purity} \\
	\midrule
	\endfirsthead 
	
	\multicolumn{6}{c}{{\small \itshape Table \thetable{} -- Continued from the previous page}} \\[0.5em]
	\toprule
	\textbf{Dataset} & \textbf{Method} & \textbf{NMI} & \textbf{ARI} & \textbf{FMI} & \textbf{Purity} \\
	\midrule
	\endhead 
	
	\midrule
	\multicolumn{6}{r}{{\small \itshape Continued on the next page...}} \\
	\endfoot
	
	\bottomrule
	\endlastfoot
	
	{} & {min}& 1.000 & 1.000 & 1.000 & 1.000\\
	\cline{2-6}
	{} & {max} & {1.000} & {1.000} & {1.000} & {1.000}\\
	\cline{2-6}
	{} & {cen} & {1.000} & {1.000} & {1.000} & {1.000}\\
	\cline{2-6}
	{} & {mean} & {1.000} & {1.000} & {1.000} & {1.000}\\
	\cline{2-6}
	{2circles} & {AChameleon} & {1.000} & {1.000} & {1.000} & {1.000}\\	
	\cline{2-6}
	{} & {RNN-DBSCAN} & {1.000} & {1.000} & {1.000} & {1.000}\\
	\cline{2-6}
	{} & {McDPC} & {1.000} & {1.000} & {1.000} & {1.000}\\
	\cline{2-6}
	{} & {G-RMS} & {0.623} & {0.645} & {0.901} & {0.902}\\
	\cline{2-6}
	{} & \textbf{\textit{\color{black}DHC-ECS}} & {\textbf{1.000}} & {\textbf{1.000}} & {\textbf{1.000}} & {\textbf{1.000}}\\
	
	\hline
	{} & {min}& 0.832 & 0.696 & 0.812 & 0.827\\
	\cline{2-6}
	{} & {max} & {0.900} & {0.760} & {0.859} & {0.957}\\
	\cline{2-6}
	{} & {cen} & {0.993} & {0.994} & {0.998} & {0.998}\\
	\cline{2-6}
	{} & {mean} & {0.846} & {0.752} & {0.817} & {0.827}\\
	\cline{2-6}
	{Aggregation} & {AChameleon} & {0.990} & {0.994} & {0.996} & {0.996}\\
	\cline{2-6}
	{} & {RNN-DBSCAN} & {0.996} & {0.998} & {0.999} & {0.999}\\
	\cline{2-6}
	{} & {McDPC} & {1.000} & {1.000} & {1.000} & {1.000}\\
	\cline{2-6}
	{} & {G-RMS} & {0.988} & {0.993} & {0.996} & {0.996}\\
	\cline{2-6}
	{} & \textbf{\textit{\color{black}DHC-ECS}} & {\textbf{0.990}} & {\textbf{0.994}} & {\textbf{0.996}} & {\textbf{0.996}}\\	
	
	\hline
	{} & {min}& 1.000 & 1.000 & 1.000 & 1.000 \\
	\cline{2-6}
	{} & {max} & {1.000} & {1.000} & {1.000} & {1.000}\\
	\cline{2-6}
	{} & {cen} & {1.000} & {1.000} & {1.000} & {1.000}\\
	\cline{2-6}
	{} & {mean} & {1.000} & {1.000} & {1.000} & {1.000}\\
	\cline{2-6}
	{Halfkernel} & {AChameleon} & {1.000} & {1.000} & {1.000} & {1.000}\\
	\cline{2-6}
	{} & {RNN-DBSCAN} & {0.743} & {0.729} & {0.884} & {1.000}\\
	\cline{2-6}
	{} & {McDPC} & {1.000} & {1.000} & {1.000} & {1.000}\\
	\cline{2-6}
	{} & {G-RMS} & {0.756} & {0.739} & {0.891} & {0.821}\\
	\cline{2-6}
	{} & \textbf{\textit{\color{black}DHC-ECS}} & \textbf{1.000} & \textbf{1.000} & \textbf{1.000} & \textbf{1.000}\\	
	
	\hline
	{} & {min}& 0.073 & -0.041 & 0.665 & 0.665 \\
	\cline{2-6}
	{} & {max} & {0.088} & {-0.045} & {0.659} & {0.638}\\
	\cline{2-6}
	{} & {cen} & {0.024} & {0.013} & {0.647} & {0.646}\\
	\cline{2-6}
	{} & {mean} & {0.161} & {0.163} & {0.671} & {0.725}\\
	\cline{2-6}
	{Flame} & {AChameleon} & {1.000} & {1.000} & {1.000} & {1.000}\\
	\cline{2-6}
	{} & {RNN-DBSCAN} & {0.935} & {0.972} & {0.992} & {0.996}\\
	\cline{2-6}
	{} & {McDPC} & {1.000} & {1.000} & {1.000} & {1.000}\\
	\cline{2-6}
	{} & {G-RMS} & {0.888} & {0.934} & {0.983} & {0.983}\\
	\cline{2-6}
	{} & \textbf{\textit{\color{black}DHC-ECS}} & \textbf{0.971} & \textbf{0.988} & \textbf{0.996} & \textbf{1.000}\\	
	
	\hline
	{} & {min}& 1.000 & 1.000 & 1.000 & 1.000\\
	\cline{2-6}
	{} & {max} & {1.000} & {1.000} & {1.000} & {1.000}\\
	\cline{2-6}
	{} & {cen} & {1.000} & {1.000} & {1.000} & {1.000}\\
	\cline{2-6}
	{} & {mean} & {1.000} & {1.000} & {1.000} & {1.000}\\
	\cline{2-6}
	{Jain} & {AChameleon} & {1.000} & {1.000} & {1.000} & {1.000}\\
	\cline{2-6}
	{} & {RNN-DBSCAN} & {1.000} & {1.000} & {1.000} & {1.000}\\
	\cline{2-6}
	{} & {McDPC} & {1.000} & {1.000} & {1.000} & {1.000}\\		
	\cline{2-6}
	{} & {G-RMS} & {0.884} & {0.945} & {0.960} & {0.260}\\
	\cline{2-6}
	{} & \textbf{\textit{\color{black}DHC-ECS}} & \textbf{1.000} & \textbf{1.000} & \textbf{1.000} & \textbf{1.000}\\	
	
	\hline
	{} & {min}& 0.253 & 0.080 & 0.550 & 0.510\\
	\cline{2-6}
	{} & {max} & {0.113} & {0.088} & {0.518} & {0.513}\\
	\cline{2-6}
	{} & {cen} & {0.539} & {0.453} & {0.681} & {0.733}\\
	\cline{2-6}
	{} & {mean} & {0.297} & {0.117} & {0.569} & {0.570}\\
	\cline{2-6}
	{Pathbased} & {AChameleon} & {0.905} & {0.930} & {0.977} & {0.977}\\
	\cline{2-6}
	{} & {RNN-DBSCAN} & {0.876} & {0.916} & {0.971} & {0.973}\\
	\cline{2-6}
	{} & {McDPC} & {1.000} & {1.000} & {1.000} & {1.000}\\
	\cline{2-6}
	{} & {G-RMS} & {0.683} & {0.635} & {0.760} & {0.190}\\
	\cline{2-6}
	{} & \textbf{\textit{\color{black}DHC-ECS}} & \textbf{0.953} & \textbf{0.969} & \textbf{0.990} & \textbf{0.990}\\
	
	\hline
	{} & {min}& 0.952 & 0.853& 0.904& 0.863\\
	\cline{2-6}
	{} & {max} & {0.923} & {0.788} & {0.852} & {0.802}\\
	\cline{2-6}
	{} & {cen} & {0.989} & {0.987} & {0.993} & {0.993}\\
	\cline{2-6}
	{} & {mean} & {0.837} & {0.427} & {0.705} & {0.668}\\
	\cline{2-6}
	{R15} & {AChameleon} & {0.992} & {0.989} & {0.995} & {0.995}\\
	\cline{2-6}
	{} & {RNN-DBSCAN} & {0.988} & {0.982} & {0.992} & {0.992}\\
	\cline{2-6}
	{} & {McDPC} & {0.977} & {0.923} & {0.952} & {0.930}\\
	\cline{2-6}
	{} & {G-RMS} & {0.993} & {0.991} & {0.996} & {0.930}\\
	\cline{2-6}
	{} & \color{black}\textbf{\textit{DHC-ECS}} & \textbf{0.984} & \textbf{0.979} & \textbf{0.990} & \textbf{0.995}\\
	
	\hline
	{} & {min}& 0.259 & 0.069 & 0.529 & 0.529\\
	\cline{2-6}
	{} & {max} & {0.259} & {0.069} & {0.509} & {0.529}\\
	\cline{2-6}
	{} & {cen} & {0.345} & {0.220} & {0.625} & {0.577}\\
	\cline{2-6}
	{} & {mean} & {0.102} & {0.009} & {0.492} & {0.397}\\
	\cline{2-6}
	{Spiral} & {AChameleon} & {1.000} & {1.000} & {1.000} & {1.000}\\
	\cline{2-6}
	{} & {RNN-DBSCAN} & {1.000} & {1.000} & {1.000} & {1.000}\\
	\cline{2-6}
	{} & {McDPC} & {1.000} & {1.000} & {1.000} & {1.000}\\
	\cline{2-6}
	{} & {G-RMS} & {1.000} & {1.000} & {1.000} & {1.000}\\
	\cline{2-6}
	{} & \textbf{\textit{\color{black}DHC-ECS}} & \textbf{1.000} & \textbf{1.000} & \textbf{1.000} & \textbf{1.000}\\		
	
	\hline
	{} & {min}& 0.568 & 0.473 & 0.831 & 0.836 \\
	\cline{2-6}
	{} & {max} & {0.342} & {0.131} & {0.651} & {0.620}\\
	\cline{2-6}
	{} & {cen} & {0.321} & {0.117} & {0.660} & {0.592}\\
	\cline{2-6}
	{} & {mean} & {0.316} & {0.112} & {0.661} & {0.661}\\
	\cline{2-6}
	{Two circles} & {AChameleon} & {1.000} & {1.000} & {1.000} & {1.000}\\
	\cline{2-6}
	{noise} & {RNN-DBSCAN} & {0.969} & {0.979} & {0.994} & {1.000}\\
	\cline{2-6}
	{} & {McDPC} & {1.000} & {1.000} & {1.000} & {1.000}\\
	\cline{2-6}
	{} & {G-RMS} & {0.734} & {0.683} & {0.808} & {0.871}\\
	\cline{2-6}
	{} & \textbf{\textit{\color{black}DHC-ECS}} & \textbf{1.000} & \textbf{1.000} & \textbf{1.000} & \textbf{1.000}\\
	
	\hline
	{} & {min}& 0.762 & 0.548 & 0.721 & 0.777 \\
	\cline{2-6}
	{} & {max} & {0.810} & {0.661} & {0.813} & {0.847}\\
	\cline{2-6}
	{} & {cen} & {0.899} & {0.856} & {0.885} & {0.882}\\
	\cline{2-6}
	{} & {mean} & {0.789} & {0.682} & {0.774} & {0.777}\\
	\cline{2-6}
	{Compound} & {AChameleon} & {0.789} & {0.607} & {0.768} & {0.855}\\
	\cline{2-6}
	{} & {RNN-DBSCAN} & {0.878} & {0.879} & {0.883} & {0.917}\\
	\cline{2-6}
	{} & {McDPC} & {0.943} & {0.973} & {0.977} & {0.980}\\
	\cline{2-6}
	{} & {G-RMS} & {0.780} & {0.691} & {0.766} & {0.018}\\
	\cline{2-6}
	{} & \textbf{\textit{\color{black}DHC-ECS}} & \textbf{0.979} & \textbf{0.990} & \textbf{0.993} & \textbf{0.993}\\
	
\end{longtable}
}

\begin{table}[!h]
	\footnotesize
	\caption{Parameter Settings of the Proposed Method Applied for the Synthetic Datasets}
	\label{table:parameters_applied}
	\renewcommand{\arraystretch}{0.9} 
	\begin{tabular}{ccccc}			
		\toprule
		\textbf{Dataset} &  $K$  & $T_\text{s}$ & $T_\text{n}$ & $r$\\
		\midrule	
		{2 circles} &10 & 0.38 & $\sqrt{N}$ & 3 \\
		{Aggregation} & 10 & 0.55 &  $\sqrt{N}$  & 3 \\
		{Halfkernel} & 10 & 0.23 &  $\sqrt{N}$ & 3 \\	
		{Flame} & 10 & 0.42 &  $\sqrt{N}$ & 3 \\
		{Jain} & 10 & 0.14 & $\sqrt{N}$  & 3 \\
		{Pathbased} & 10 & {0.38} &  $\sqrt{N}$  & 3 \\
		{R15} & 10 & 0.38 &  $\sqrt{N}$ & 3 \\
		{Spiral} & 8 & 0.38 &  $\sqrt{N}$ & 3 \\
		{Two circles noise}& 10 & 0.43 & $\sqrt{N}$ & 3\\
		{Compound} & 7 & 0.51 & 0.95$\sqrt{N}$  & 3\\
		\bottomrule
		\vspace{-7mm}
	\end{tabular}
\end{table}	

\subsection{Computational Complexity}

The computational complexity of the proposed DHC-ECS algorithm is mainly determined by the hierarchical agglomerative merging procedure.

In the first stage, the pairwise distance matrix is first constructed, requiring $O(N^2)$ time for a dataset containing $N$ observations. The similarity matrix is then initialized based on the average inter-cluster distance. During each agglomerative iteration, the pair of clusters with the highest similarity is merged, followed by updating the similarities between the newly generated cluster and the remaining clusters. Since there are initially $N$ sub-clusters and approximately $N-T_{\mathrm{n}}$ merging operations are performed, where $T_{\mathrm{n}}$ denotes the predefined number of sub-clusters after the first stage, the overall computational complexity of the first stage is $O(N^3)$ under a straightforward implementation, which is consistent with the naive hierarchical clustering and can be reduced to $O(N^2\log N)$ through a priority-queue implementation as discussed in \cite{schutze2008introduction}.

In the second stage, the KNN connection subgraphs are first constructed, requiring $O(T_{\mathrm{n}}^2)$ time using exhaustive nearest-neighbor search (or $O(T_{\mathrm{n}}\log T_{\mathrm{n}})$ for low-dimensional datasets when efficient nearest-neighbor search algorithms are adopted). Subsequently, the KNN connection subgraph is extracted for each neighboring cluster pair, and the proposed similarity indices, including LC, LS, DS, and CV, are computed only for cluster pairs connected by the ECS. Assuming that the average number of neighboring clusters remains bounded, the number of candidate cluster pairs at the $t$-th merging iteration is approximately $O(N^t)$, where $N^t$ denotes the current number of clusters. Furthermore, since the number of preliminary sub-clusters has already been reduced to $T_{\mathrm{n}}$ ($T_{\mathrm{n}}\ll N$, typically $T_{\mathrm{n}}\approx\sqrt{N}$), and the second-stage merging terminates at a much smaller number of clusters $L$, the overall complexity of the second-stage merging procedure can be approximated as $O(T_{\mathrm{n}}^2)$ given that $L$ is typically much smaller than $T_{\mathrm{n}}$. Therefore, the complexity of the second-stage agglomerative merging is significantly smaller than that of the first stage, thus the overall worst-case computational complexity of DHC-ECS is dominated by the first-stage hierarchical merging and is given by $	O(N^3)+O(T_{\mathrm{n}}^2)\approx O(N^3)$. The space complexity is mainly determined by the storage of the pairwise distance, similarity matrix, and the KNN graph, resulting in an overall space complexity of $O(N^2)$.

Compared with representative density-based clustering algorithms, such as RNN-DBSCAN, McDPC, and G-RMS, whose computational complexities are typically around $O(N^2)$\cite{8240674,2020MATLAB,9698033}, the proposed DHC-ECS incurs a higher worst-case computational cost because of the iterative hierarchical merging procedure. Nevertheless, unlike conventional hierarchical clustering algorithms that evaluate all cluster pairs during each merging iteration, DHC-ECS computes the proposed similarity only for cluster pairs connected through the ECS. Consequently, the practical computational cost of the second-stage is usually much lower than that of exhaustive hierarchical merging, especially for sparse neighborhood graphs. In return for the additional computational cost, DHC-ECS provides a substantially richer characterization of inter-cluster relationships by jointly exploiting graph connectivity, local density, and structural information within KNN connection regions, leading to improved robustness on datasets with heterogeneous densities and complex cluster structures.

\vspace{-1mm}
\section{Conclusion}\label{sec:conclusion}

In this paper, a novel two-stage density-aware hierarchical clustering algorithm based on element-categorized KNN connection subgraphs is proposed. In the first stage, we construct preliminary clusters using a greedy algorithm. In the second stage, the ECS is proposed to represent the connections between sub-clusters, and a new similarity metric is formulated, considering not only the point connectivity but also the density information. The proposed DHC-ECS algorithm is evaluated on ten widely-used synthetic datasets with heterogeneous distributions. Comparative analyses are conducted against hierarchical clustering methods based on traditional similarity metrics, as well as several representative baseline clustering algorithms, including an improved hierarchical clustering algorithm (ACHAMELEON), the density-based algorithms RNN-DBSCAN, McDPC, and a robust mean-shift algorithm (G-RMS). For quantitative comparisons, four external evaluation indices, including NMI, ARI, FMI and Purity, are calculated. Experimental results demonstrate that the proposed algorithm exhibits superior robustness and performs well across all these datasets compared to the aforementioned algorithms. This highlights the effectiveness and applicability of the DHC-ECS algorithm to obtain more refined clustering results across datasets with different features, including convex and non-convex shapes, uniform and non-uniform densities, balanced or unbalanced distributions, single-centered or multi-centered structures. Future work includes conducting extended experiments on high-dimensional datasets, and searching for improvements on computational complexity to accommodate to larger datasets. 






\vspace{-1mm}
\bibliographystyle{unsrtnat}
\bibliography{Bibliography}

\begin{thebibliography}{36}
\providecommand{\natexlab}[1]{#1}
\providecommand{\url}[1]{\texttt{#1}}
\expandafter\ifx\csname urlstyle\endcsname\relax
  \providecommand{\doi}[1]{doi: #1}\else
  \providecommand{\doi}{doi: \begingroup \urlstyle{rm}\Url}\fi

\bibitem[Bryant and Cios(2018)]{8240674}
Avory Bryant and Krzysztof Cios.
\newblock {RNN-DBSCAN}: A density-based clustering algorithm using reverse
  nearest neighbor density estimates.
\newblock \emph{IEEE Transactions on Knowledge and Data Engineering},
  30\penalty0 (6):\penalty0 1109--1121, 2018.
\newblock \doi{10.1109/TKDE.2017.2787640}.

\bibitem[Khan et~al.(2014)Khan, Rehman, Aziz, Fong, and Sarasvady]{6814687}
Kamran Khan, Saif~Ur Rehman, Kamran Aziz, Simon Fong, and S.~Sarasvady.
\newblock {DBSCAN}: Past, present and future.
\newblock In \emph{The Fifth International Conference on the Applications of
  Digital Information and Web Technologies (ICADIWT 2014)}, pages 232--238,
  2014.
\newblock \doi{10.1109/ICADIWT.2014.6814687}.

\bibitem[Guha et~al.(1998)Guha, Rastogi, and Shim]{1998CURE}
Sudipto Guha, Rajeev Rastogi, and Kyuseok Shim.
\newblock {CURE}: An efficient clustering algorithm for large databases.
\newblock \emph{ACM Sigmod record}, 27\penalty0 (2):\penalty0 73--84, 1998.
\newblock \doi{10.1145/276305.276312}.

\bibitem[Guha et~al.(2000)Guha, Rastogi, and Shim]{754967}
Sudipto Guha, Rajeev Rastogi, and Kyuseok Shim.
\newblock {ROCK}: A robust clustering algorithm for categorical attributes.
\newblock \emph{Information systems}, 25\penalty0 (5):\penalty0 345--366, 2000.
\newblock \doi{10.1109/ICDE.1999.754967}.

\bibitem[Karypis et~al.(1999)Karypis, Han, and Kumar]{781637}
George Karypis, Eui-Hong Han, and Vipin Kumar.
\newblock Chameleon: Hierarchical clustering using dynamic modeling.
\newblock \emph{computer}, 32\penalty0 (8):\penalty0 68--75, 1999.
\newblock \doi{10.1109/2.781637}.

\bibitem[Nazari et~al.(2015)Nazari, Kang, Asharif, Sung, and Ogawa]{7439517}
Zahra Nazari, Dongshik Kang, M.~Reza Asharif, Yulwan Sung, and Seiji Ogawa.
\newblock A new hierarchical clustering algorithm.
\newblock In \emph{2015 International Conference on Intelligent Informatics and
  Biomedical Sciences (ICIIBMS)}, pages 148--152, 2015.
\newblock \doi{10.1109/ICIIBMS.2015.7439517}.

\bibitem[Nazari and Kang(2018)]{8596795}
Zahra Nazari and Dongshik Kang.
\newblock A new hierarchical clustering algorithm with intersection points.
\newblock In \emph{2018 5th IEEE Uttar Pradesh Section International Conference
  on Electrical, Electronics and Computer Engineering (UPCON)}, pages 1--5,
  2018.
\newblock \doi{10.1109/UPCON.2018.8596795}.

\bibitem[Guo et~al.(2019{\natexlab{a}})Guo, Zhao, and Liu]{2019Research}
Dongwei Guo, Jingjing Zhao, and Jici Liu.
\newblock Research and application of improved {CHAMELEON} algorithm based on
  condensed hierarchical clustering method.
\newblock In \emph{Proceedings of the 2019 8th international conference on
  networks, communication and computing}, pages 14--18, 2019{\natexlab{a}}.

\bibitem[Cao et~al.(2018)Cao, Su, Wang, Wang, Lv, and Li]{2018An}
Xiaoxiao Cao, Tianyun Su, Pengyu Wang, Guoyu Wang, Zhihan Lv, and Xinfang Li.
\newblock An optimized {CHAMELEON} algorithm based on local features.
\newblock In \emph{Proceedings of the 2018 10th International Conference on
  Machine Learning and Computing}, pages 184--192, 2018.

\bibitem[Ester et~al.(1996)Ester, Kriegel, Sander, and Xu]{ester1996density}
Martin Ester, Hans-Peter Kriegel, J{\"o}rg Sander, and Xiaowei Xu.
\newblock Density-based spatial clustering of applications with noise.
\newblock In \emph{Int. Conf. knowledge discovery and data mining}, volume 240,
  1996.

\bibitem[Rodriguez and Laio(2014)]{rodriguez2014clustering}
Alex Rodriguez and Alessandro Laio.
\newblock Clustering by fast search and find of density peaks.
\newblock \emph{science}, 344\penalty0 (6191):\penalty0 1492--1496, 2014.

\bibitem[Guo et~al.(2024)Guo, Qin, Cai, and Su]{guo2024hybrid}
Limin Guo, Weijia Qin, Zhi Cai, and Xing Su.
\newblock Hybrid clustering algorithm based on improved density peak
  clustering.
\newblock \emph{Applied Sciences}, 14\penalty0 (2):\penalty0 715, 2024.

\bibitem[Comaniciu and Meer(1999)]{comaniciu1999mean}
Dorin Comaniciu and Peter Meer.
\newblock Mean shift analysis and applications.
\newblock In \emph{Proceedings of the seventh IEEE international conference on
  computer vision}, volume~2, pages 1197--1203. IEEE, 1999.

\bibitem[Campello et~al.(2015)Campello, Moulavi, Zimek, and
  Sander]{campello2015hierarchical}
Ricardo~JGB Campello, Davoud Moulavi, Arthur Zimek, and J{\"o}rg Sander.
\newblock Hierarchical density estimates for data clustering, visualization,
  and outlier detection.
\newblock \emph{ACM Transactions on Knowledge Discovery from Data (TKDD)},
  10\penalty0 (1):\penalty0 1--51, 2015.

\bibitem[Xu et~al.(2016)Xu, Wang, and Deng]{xu2016denpehc}
Ji~Xu, Guoyin Wang, and Weihui Deng.
\newblock {DenPEHC}: Density peak based efficient hierarchical clustering.
\newblock \emph{Information Sciences}, 373:\penalty0 200--218, 2016.

\bibitem[Neto et~al.(2019)Neto, Sander, Campello, and
  Nascimento]{neto2019efficient}
Antonio Cavalcante~Araujo Neto, J{\"o}rg Sander, Ricardo~JGB Campello, and
  Mario~A Nascimento.
\newblock Efficient computation and visualization of multiple density-based
  clustering hierarchies.
\newblock \emph{IEEE Transactions on Knowledge and Data Engineering},
  33\penalty0 (8):\penalty0 3075--3089, 2019.

\bibitem[Zhu et~al.(2022)Zhu, Ting, Jin, and Angelova]{zhu2022hierarchical}
Ye~Zhu, Kai~Ming Ting, Yuan Jin, and Maia Angelova.
\newblock Hierarchical clustering that takes advantage of both density-peak and
  density-connectivity.
\newblock \emph{Information Systems}, 103:\penalty0 101871, 2022.

\bibitem[Wang and Li(2025)]{wang2025fast}
Renmin Wang and Jie Li.
\newblock Fast sparse representative tree splitting via local density for
  large-scale clustering.
\newblock \emph{Scientific Reports}, 15\penalty0 (1):\penalty0 29398, 2025.

\bibitem[Shao et~al.(2018)Shao, Yang, Zhang, Liu, and Kramer]{shao2018graph}
Junming Shao, Qinli Yang, Zhong Zhang, Jinhu Liu, and Stefan Kramer.
\newblock Graph clustering with local density-cut.
\newblock In \emph{International Conference on Database Systems for Advanced
  Applications}, pages 187--202. Springer, 2018.

\bibitem[Du et~al.(2024)Du, Li, Li, and Yu]{du2024adpscan}
Xinyu Du, Fangfang Li, Xiaohua Li, and Ge~Yu.
\newblock {ADPSCAN}: Structural graph clustering with adaptive density peak
  selection and noise re-clustering.
\newblock \emph{Applied Sciences}, 14\penalty0 (15):\penalty0 6660, 2024.

\bibitem[Hou et~al.(2016)Hou, Gao, and Li]{hou2016dsets}
Jian Hou, Huijun Gao, and Xuelong Li.
\newblock {DSets-DBSCAN}: A parameter-free clustering algorithm.
\newblock \emph{IEEE Transactions on Image Processing}, 25\penalty0
  (7):\penalty0 3182--3193, 2016.

\bibitem[Li et~al.(2020)Li, Zhang, Wang, and Nie]{li2020multiview}
Xuelong Li, Han Zhang, Rong Wang, and Feiping Nie.
\newblock Multiview clustering: A scalable and parameter-free bipartite graph
  fusion method.
\newblock \emph{IEEE Transactions on Pattern Analysis and Machine
  Intelligence}, 44\penalty0 (1):\penalty0 330--344, 2020.

\bibitem[Mahon and Lapata(2025)]{mahon2025k}
Louis Mahon and Mirella Lapata.
\newblock K*-means: A parameter-free clustering algorithm.
\newblock \emph{arXiv preprint arXiv:2505.11904}, 2025.

\bibitem[Ruspini et~al.(2019)Ruspini, Bezdek, and Keller]{8610270}
Enrique~H. Ruspini, James~C. Bezdek, and James~M. Keller.
\newblock Fuzzy clustering: A historical perspective.
\newblock \emph{IEEE Computational Intelligence Magazine}, 14\penalty0
  (1):\penalty0 45--55, 2019.
\newblock \doi{10.1109/MCI.2018.2881643}.

\bibitem[Guo et~al.(2019{\natexlab{b}})Guo, Zhao, and Liu]{guo2019research}
Dongwei Guo, Jingjing Zhao, and Jici Liu.
\newblock Research and application of improved {CHAMELEON} algorithm based on
  condensed hierarchical clustering method.
\newblock In \emph{Proceedings of the 2019 8th international conference on
  networks, communication and computing}, pages 14--18, 2019{\natexlab{b}}.

\bibitem[Barton et~al.(2019)Barton, Bruna, and Kordik]{barton2019chameleon}
Tomas Barton, Tomas Bruna, and Pavel Kordik.
\newblock Chameleon 2: an improved graph-based clustering algorithm.
\newblock \emph{ACM Transactions on Knowledge Discovery from Data (TKDD)},
  13\penalty0 (1):\penalty0 1--27, 2019.

\bibitem[Jeong et~al.(2013)Jeong, Yoon, Song, Lee, Ryu, Kim, and
  Jeong]{jeong2013data}
Kyo-Sung Jeong, Seok-Ho Yoon, Suk-Soon Song, Sang-Chul Lee, Minsoo Ryu,
  Sang-Wook Kim, and Byung-Soo Jeong.
\newblock Data partitioning in hierarchical clustering: A parameter-insensitive
  approach.
\newblock \emph{International Information Institute (Tokyo). Information},
  16\penalty0 (10):\penalty0 7699, 2013.

\bibitem[Singh and Ahuja(2025)]{singh2025chameleon2++}
Priyanshu Singh and Kapil Ahuja.
\newblock Chameleon2++: An efficient and scalable variant of chameleon
  clustering.
\newblock \emph{arXiv preprint arXiv:2501.02612}, 2025.

\bibitem[Fr\"anti and Sieranoja(2018)]{ClusteringDatasets}
Pasi Fr\"anti and Sami Sieranoja.
\newblock K-means properties on six clustering benchmark datasets, 2018.
\newblock URL \url{http://cs.uef.fi/sipu/datasets/}.

\bibitem[Wang et~al.(2020)Wang, Wang, Zhang, Pang, and Zhou]{2020MATLAB}
Yizhang Wang, Di~Wang, Xiaofeng Zhang, Wei Pang, and You Zhou.
\newblock {McDPC}: multi-center density peak clustering.
\newblock \emph{Neural Computing and Applications}, 32\penalty0 (17), 2020.
\newblock \doi{10.1007/s00521-020-04754-5}.

\bibitem[Cariou et~al.(2022)Cariou, Le~Moan, and Chehdi]{9698033}
Claude Cariou, Steven Le~Moan, and Kacem Chehdi.
\newblock A novel mean-shift algorithm for data clustering.
\newblock \emph{IEEE Access}, 10:\penalty0 14575--14585, 2022.
\newblock \doi{10.1109/ACCESS.2022.3147951}.

\bibitem[Manning(2009)]{2008Introduction}
Christopher~D Manning.
\newblock \emph{An introduction to information retrieval}.
\newblock 2009.

\bibitem[Vinh et~al.(2010)Vinh, Epps, and Bailey]{0Information}
Nguyen~Xuan Vinh, Julien Epps, and James Bailey.
\newblock Information theoretic measures for clusterings comparison: Variants,
  properties, normalization and correction for chance.
\newblock \emph{Journal of Machine Learning Research}, 11:\penalty0 2837--2854,
  2010.

\bibitem[Powers(2011)]{2011Powers}
David~MW Powers.
\newblock Evaluation: from precision, recall and {F-measure} to {ROC},
  informedness, markedness and correlation.
\newblock \emph{Jonunal of Machine Learning Technologies}, 2:\penalty0
  2229--3981, 2011.

\bibitem[Beyer et~al.(1999)Beyer, Goldstein, Ramakrishnan, and
  Shaft]{beyer1999nearest}
Kevin Beyer, Jonathan Goldstein, Raghu Ramakrishnan, and Uri Shaft.
\newblock When is ``nearest neighbor'' meaningful?
\newblock In \emph{Database Theory—ICDT’99: 7th International Conference
  Jerusalem, Israel, January 10--12, 1999 Proceedings 7}, pages 217--235.
  Springer, 1999.

\bibitem[Sch{\"u}tze et~al.(2008)Sch{\"u}tze, Manning, and
  Raghavan]{schutze2008introduction}
Hinrich Sch{\"u}tze, Christopher~D Manning, and Prabhakar Raghavan.
\newblock \emph{Introduction to information retrieval}, volume~39.
\newblock Cambridge University Press Cambridge, 2008.

\end{thebibliography}

\appendix

\end{document}